\documentclass{article}

\usepackage[preprint]{neurips_2024}
\usepackage{natbib}
\setcitestyle{numbers}

\usepackage[utf8]{inputenc} 
\usepackage[T1]{fontenc}    
\usepackage{hyperref}       
\usepackage{url}            
\usepackage{booktabs}       
\usepackage{amsfonts}       
\usepackage{microtype}      
\usepackage{xcolor}         
\usepackage{multirow}
\usepackage{algpseudocode}
\usepackage{algorithm} 
\usepackage{listings} 

\makeatother

\usepackage{comment}
\usepackage{lipsum}
\usepackage{array}

\usepackage[utf8]{inputenc}

\usepackage{microtype}

\usepackage{inconsolata}

\usepackage{graphicx}
\usepackage{tabularx} 

\usepackage{amsmath}
\usepackage{amssymb}
\usepackage{booktabs}
\usepackage{makecell}

\usepackage{color, booktabs,subcaption,amsfonts,dcolumn}
\usepackage{xcolor}
\usepackage{mdframed}
\usepackage{subcaption}
\newmdenv[leftmargin=5pt,rightmargin=5pt,backgroundcolor=gray!20,innertopmargin=5pt,innerbottommargin=5pt]{coloredquote}

\usepackage{amsfonts}
\usepackage{graphicx}

\usepackage{multirow,diagbox,makecell}
\usepackage{tikz}

\usepackage{pifont}
\usepackage{booktabs}

\usepackage{capt-of}

\title{Smart Language Agents in Real-World Planning}

\author{
  Annabelle Miin$^*$, Timothy Wei$^*$\\
  Pacific Collegiate School, Saratoga High School\\
  \texttt{annabellemiin8@gmail.com, timswei@gmail.com} \\
}

\begin{document}

\maketitle

\def\thefootnote{*}\footnotetext{These authors contributed equally to this work}
\def\thefootnote{}\footnotetext{GitHub Link: \url{https://github.com/llv22/TravelPlanner_forward}}

\begin{abstract}
Comprehensive planning agents have been a long term goal in the field of artificial intelligence. Recent innovations in Natural Language Processing have yielded success through the advent of Large Language Models (LLMs). We seek to improve the travel-planning capability of such LLMs by extending upon the work of the previous paper TravelPlanner (Xie et al., \cite{xie2024travelplanner}). Our objective is to explore a new method of using LLMs to improve the travel planning experience. We focus specifically on the “sole-planning” mode of travel planning; that is, the agent is given necessary reference information, and its goal is to create a comprehensive plan from the reference information. While this does not simulate the real-world we feel that an optimization of the sole-planning capability of a travel planning agent will still be able to enhance the overall user experience. We propose a semi-automated prompt generation framework which combines the LLM-automated prompt and "human-in-the-loop" to iteratively refine the prompt to improve the LLM performance. Our result shows that LLM automated prompt has its limitations and "human-in-the-loop" greatly improves the performance by 139\% with one single iteration.

\end{abstract}

\section{Introduction}
The history of artificial intelligence (AI) and large language models (LLMs) in constraint-based planning dates back to the early days of AI research. Constraint satisfaction problems (CSPs) have been utilized for various applications including scheduling, resource allocation, and problem-solving. Early works by researchers such as Dechter and Pearl \cite{dechter1988ai} laid the foundation for understanding the theoretical underpinnings of CSPs and developed algorithms for efficient problem-solving within these frameworks. More recently LLMs have significantly enhanced the capability and flexibility of constraint-based systems.

Large language models, such as GPT-3 and its successors, have revolutionized the field by enabling more natural and flexible interactions with constraint-based planning systems. Recent studies by Brown et al. \cite{brown2020learners} and Bommasani et al. \cite{rishi2021foundation} have demonstrated the potential of these models ranging from natural language understanding to decision-making support systems. The integration of LLMs with constraint-based planning can better accommodate complex user requirements and preferences.

Our work focuses on enhancing the ability of LLMs to achieve goals under constraints. This includes systematically analyzing current benchmarks, proposing a general framework for improving LLMs, automating constraint understanding based on system resources and implicit human feedback, and improving the travel planner benchmark. Our advancements demonstrate the potential for achieving human-level performance with LLM automated prompt plus minimal human intervention.

\section{Related work}
A multi-constraint problem can be challenging for humans to solve, let alone a large language model. Developing a framework around the types of constraints and categorizing them into key basic types dramatically simplifies the process of decomposing the problem and the data needed to train, validate and test the model. It is useful to bucket the constraints into 1) \textbf{environment constraints} or constantly changing inputs due to real world conditions such as flight availability on certain dates 2) \textbf{common sense constraints} or what a reasonable person might deem to make sense such as limiting driving to 8 or 10 hours a day and 3) \textbf{hard constraints} or well defined requirements such as a maximum budget (Xie et al.,\cite{xie2024travelplanner}). 

It also makes it easier to set and evaluate pass and fail criteria by bucketing. While this evaluation (and prompt improvement) may also be automated, getting to the correct or acceptable solution purely through automated iterative experimentation can be very time consuming and inefficient. Using a "human-in-the-loop" system (Xin et al., \cite{xin2018hitl}) has the potential to dramatically speed things up, sometimes up to 10x, versus typical iterative workflows. Whereas Xin et al., does intermediate results reuse and introspection between iterations, automates background search during think time, and automates cues for iterative changes, we take a different approach by using human failure analysis to identify model weaknesses and use that to directly improve the automatically generated prompt.

\section{Methods: Framework for improving LLMs}


\begin{figure}[H]
    \centering
    \includegraphics[width=\linewidth]{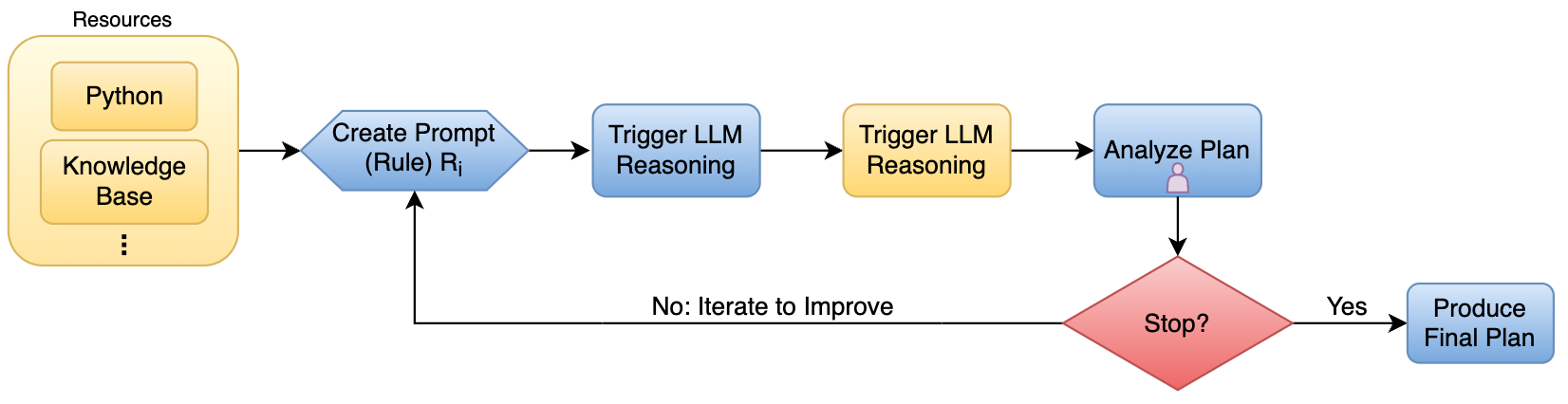}
    \caption{Framework for using automated prompt and "human-in-the-loop" iteration to improve LLM capability to produce final plan. Here, resources at the starting point, reference to artifacts that contain the constraints that the Planner needs to conform to. We will improve the prompt via "human-in-the-loop" iteration and $R_i$ stands for the generated prompt at $i$th iteration. During each iteration, the generated prompt will be used for LLM reasoning to produce a plan.  When the performance from $R_{i-1}$ and $R_i$ are very close with each other, then the iteration stops. All data in the system are colored in yellow. All activities are colored in Blue.}
    \label{fig:framework}
\end{figure}

 In this paper, we propose a general framework for creating an effective prompt for LLMs and apply it to the travel planning use case \cite{xie2024travelplanner} where a travel plan is generated with the help of LLMs. The framework consists of two main steps as depicted in Figure \ref{fig:framework}. The $1st$ step is creating an initial prompt by LLMs through its automatic summarization of external resources such as python evaluation script and reference data in our case. This automatic step reduces labor intensity when creating the initial prompt for use. The $2nd$ step is further improving the prompt through prompt-tuning with human-in-the-loop where manual failure analysis identifies failed plans in the training data and addresses agent faults in the prompt such as adding a corrected plan for the failed plan to the prompt. This $2nd$ step is an iterative process to optimize the prompt using the training data to generate prompt $R_i$ at iteration $i$. The iteration can stop when the performance converges between consecutive iterations.
 
 The initial automated prompt mainly covers the extracted rule and the improved prompts focuses on picking up the tricky failure cases, but it is still an open question whether just simply concatenating them is the right way to make LLM yield better result. Humans can analyze the extracted rules and failure examples and re-organize or condense the prompt to avoid some side effects, like lengthy content, etc. For that reason, we also manually created a prompt and use its performance to serve as a gold standard for comparison. It gives us the opportunity to measure the gap in performance between the semi-automated prompt and the manually created prompt. In addition, we made improvement by using structured reference information rather than unstructured information given to a travel-planning agent.

\section{Experiments}
\subsection{Set-up}
The set-up for our experiments consisted of selecting an LLM to use to automatically generate the constraints in step 1 and selecting an LLM to use for the travel-planning agent. We decided to use GPT-4 Turbo as the base travel-planning agent model due to its high performance as the state-of-the-art language model as well as its easy accessibility through its API. We chose to utilize GPT-4o to automate the generation of the constraints due to it being a cheaper option than GPT-4 Turbo given that there is a free quota every day. Additionally, by utilizing the GPT models, we used the OpenAI API that outsources the computing hardware required.

\subsection{Structured Reference Information}
Previous papers' trials of the travel-planning agent utilized reference information in the form of raw JSON \cite{xie2024travelplanner}. While the JSON itself is structured, the information contained within each JSON key was actually in CSV form, so we felt that restructuring the reference information as a series of CSV files would allow the travel-planning agent to perform better. To this end, we created a script that converted each raw JSON reference information into a list of CSV files. Then, we appended this structured reference information to each prompt to achieve better results.

\subsection{GPT-4 Turbo Baseline}
To obtain a baseline to compare our results to, we utilized the TravelPlanner \cite{xie2024travelplanner} paper's Github repository, utilizing its evaluation scripts and plan generation datasets. We call these results the \textit{\textbf{GPT-4 Turbo Baseline}}. Additionally, we used their zero-shot prompt, with unstructured reference data, as a baseline to compare to.

\subsection{Automation Module}
To automate the prompt rule generation, we used GPT-4o to summarize the evaluation script's rules and created a prompt without examples to avoid mismatches with the evaluation criteria, which we call the \textit{\textbf{Initial GPT-4o Generated Prompt}}. To further enhance the prompt, we added a manually selected example from the TravelPlanner training dataset's most challenging plans, resulting in a prompt combining automated rules and a curated example. This result is called the \textit{\textbf{Improved GPT-4o Prompt after 1st iteration}}. While we only ran this automation module loop once, we believe that further loops would help increase performance further.

\subsection{Manual Prompt Creation}
After analyzing the GPT-4 Turbo Baseline results, we identified and included poorly performing constraints in the prompt, leading to significant performance improvements. This resulted in the "gold standard" for our automation module to aspire to, which we call the \textit{\textbf{GPT-4 Turbo Manually Modified Prompt}}.

\subsection{Evaluation}
To evaluate our travel-planning agents, we utilize the TravelPlanner paper's evaluation algorithm. This algorithm evaluates the plan generated by the TravelPlanner using a number of constraints, split between \textit{\textbf{Commonsense Constraints}} and \textit{\textbf{Hard Constraints}}. Commonsense Constraints consist of those constraints that a traveler is not required to make, but will naturally make. For example, one commonsense constraint could be to require Diverse Attractions, as it would not make sense for multiple days to visit the same attraction. Hard Constraints, on the other hand, are things that human travelers are required to follow. These include constraints such as the budget for the trip as well as abiding by all the room rules of each accommodation that is booked. 

\subsection{Data Splits}
We utilize the data splits from TravelPlanner \cite{xie2024travelplanner}, and these splits can be obtained in our GitHub. Specifically, to evaluate our results, we utilize the validation split with 180 queries, and to manually fine-tune as well as fine-tune the automated prompt, we utilize the 45-query training split.

\section{Results}

Table \ref{main_result} below shows the passing rates. Row "GPT-4 Turbo Baseline" shows the result from TravelPlanner \cite{xie2024travelplanner}. The next three rows contain the results from manually modified prompt, initial GPT-4o automated prompt and the improved prompt after 1st iteration with "human-in-the-loop".

\begin{table*}[h]
\centering
\caption{Agent Performance on Validation Data}
\label{main_result}
\setlength{\abovecaptionskip}{0.2cm}
\setlength\tabcolsep{3pt}
\resizebox{\textwidth}{!}{
\begin{tabular}{l|c|cc|cc|c} 
\toprule
           & \multirow{2}{*}{\textbf{Delivery Rate}} & \multicolumn{2}{c|}{\textbf{Commonsense Pass Rate~(\%)}} & \multicolumn{2}{c|}{\textbf{Hard Constraint Pass Rate~(\%)}} & \multirow{2}{*}{\textbf{Final Pass Rate}}  \\ 
\cline{3-6}
           &                                      & \textbf{Micro} & \textbf{Macro} & \textbf{Micro} & \textbf{Macro}\\ 
\hline
GPT-4 Turbo Baseline & 100.0 & 80.07 & 17.78 & 50.23 & 28.33 & 5.55 \\
GPT-4 Turbo Manually Modified Prompt & 100.0 & \textbf{84.93} & \textbf{27.22} & \textbf{61.19} & \textbf{42.78} & \textbf{12.78} \\
Initial GPT-4o Generated Prompt & 100.0 & 81.39 & 15.56 & 37.14 & 22.22 & 2.78 \\
Improved GPT-4o Prompt after 1st Iteration & 100.0 & 81.39 & 18.33 \textbf{(+2.8)} & 54.76 \textbf{(+17.62)} & 34.44 \textbf{(+12.22)} & 6.67 \textbf{(+3.89)}\\
\bottomrule
\end{tabular}}
\end{table*}

From Table \ref{main_result}, we can see that the initial GPT-4o automatically generated prompt produces unsatisfactory result, lower than the GPT-4 Turbo Baseline. However, with one iteration of "human-in-the-loop" to augment the automated prompt, the success rate of the GPT-4o generated prompt significantly improves, with the final pass rate increasing from 2.78 to 6.67 (139\% improvement). The numbers in the parenthesis in the last row show the increases from one iteration. The resulting performance metrics are now greater than the GPT-4 Turbo Baseline. In the meanwhile, the GPT-4 Manually Modified Prompt performs much better than the GPT-4 Turbo Baseline, achieving greater pass rates across the board as well as a final pass rate that is 2.3 times that of the baseline (12.78 vs. 5.55). With one single iteration of "human-in-the-loop", we are getting closer to the performance of the GPT-4 Manually Modified Prompt and we are hopeful the gaps will become smaller with more iterations.


\section{Limitations}
Our framework relies mostly on automation to generate prompts and evaluate the results. It also uses human-in-the-loop as the feedback mechanism. There are a few key limitations to our framework. First, the reference data in \cite{xie2024travelplanner} is limited in terms of the selection of travel information, accommodations, restaurants and attractions. We didn't have sufficient time to include a more diverse set of reference data for the agent to use. Second, we used our own evaluation script to assess the efficacy of the resultant LLM generated travel plans; however, LLMs don't inherently have this capability built-in. Third, while we have shown that this framework is applicable to travel planning, we have not tested it with other use cases (i.e. such as planning a high school or college student's multi-year course load given budgetary, degree requirements and other constraints). Finally, human-in-the-loop is inherently not scalable due to the manual nature of the analysis, and another LLM agent should be designed for alleviating this drawback. 


\section{Conclusion}
We developed a framework of using LLMs with a "human-in-the-loop" feedback mechanism to automate and solve constraint-based problems. We evaluated this framework with the travel planning problem \cite{xie2024travelplanner}. The LLM-automated prompt has poor performance. We are able to improve the pass rate by 139\% by one single "human-in-the-loop" iteration which produces result on-par with the baseline and closer to the human manually created prompt. While there are limitations with this framework, we believe that this framework has great potential to achieve more accurate results and has broader applicability to solve constraint-based problems which make it a worthwhile endeavor. Aside from travel planning, the "human-in-the-loop" approach has far-reaching potentials such as enhancing image recognition for wildfire detection and optimizing wearable monitoring devices for more precise heart rate and SpO2 readings. The possibilities are truly endless; this novel approach to improve LLM promises to simplify decision-making across various industries, leading to a more efficient and equitable future where data-driven insights benefit everyone.

\begin{ack}
We appreciate the support and guidance from our mentor Lei Ding, a resident LLM expert and a PhD student from University of California at Santa Cruz. We did not receive any outside funding and do not have any competing interests.

\end{ack}

\bibliographystyle{unsrtnat}
\bibliography{neurips_2024}

\begin{thebibliography}{5}
\providecommand{\natexlab}[1]{#1}
\providecommand{\url}[1]{\texttt{#1}}
\expandafter\ifx\csname urlstyle\endcsname\relax
  \providecommand{\doi}[1]{doi: #1}\else
  \providecommand{\doi}{doi: \begingroup \urlstyle{rm}\Url}\fi

\bibitem[Xie et~al.(2024)Xie, Zhang, Chen, Zhu, Lou, Tian, Xiao, and Su]{xie2024travelplanner}
Jian Xie, Kai Zhang, Jiangjie Chen, Tinghui Zhu, Renze Lou, Yuandong Tian, Yanghua Xiao, and Yu~Su.
\newblock Travelplanner: A benchmark for real-world planning with language agents.
\newblock \emph{arXiv preprint arXiv:2402.01622}, 2024.

\bibitem[Dechter and Pearl(1988)]{dechter1988ai}
Rina Dechter and Judea Pearl.
\newblock Network-based heuristics for constraint-satisfaction problems.
\newblock \emph{Artificial Intelligence}, 1988.

\bibitem[Brown et~al.(2020)Brown, Mann, Ryder, Subbiah, Kaplan, Dhariwal, Neelakatan, Shyam, Sastry, Askell, Agarwal, Herbert-Voss, Krueger, Henighan, Child, Ramesh, Ziegler, We, Winter, Hesee, Chen, Sigler, Litwin, Gray, Chess, Clark, Berner, McCandlish, Radford, Sutskever, and Amodei]{brown2020learners}
Tom Brown, Benjamin Mann, Nick Ryder, Melanie Subbiah, Jared~D. Kaplan, Prafulla Dhariwal, Arvind Neelakatan, Pranav Shyam, Girish Sastry, Amanda Askell, Sandhini Agarwal, Ariel Herbert-Voss, Gretchen Krueger, Tom Henighan, Rewon Child, Aditya Ramesh, Daniel Ziegler, Jeffrey We, Clemens Winter, Chris Hesee, Mark Chen, Eric Sigler, Mateusz Litwin, Scott Gray, Benjamin Chess, Jack Clark, Christopher Berner, Sam McCandlish, Alec Radford, Ilya Sutskever, and Dario Amodei.
\newblock Language models are few-shot learners.
\newblock \emph{Advances in Neural Information Processing Systems 33}, 2020.

\bibitem[Bommasani et~al.(2021)Bommasani, Hudson, and et. al]{rishi2021foundation}
Rishi Bommasani, Drew~A. Hudson, and et. al.
\newblock On the opportunities and risks of foundation models.
\newblock \emph{arXiv preprint arXiv:2108.07258}, 2021.

\bibitem[Xin et~al.(2018)Xin, Ma, Liu, Macke, Song, and Parameswaran]{xin2018hitl}
Doris Xin, Litian Ma, Jialin Liu, Stephen Macke, Schuchen Song, and Aditya Parameswaran.
\newblock Accelerating human-in-the-loop machine learning: Challenges and opportunities.
\newblock \emph{arXiv preprint arXiv:1804.05892}, 2018.

\end{thebibliography}

\end{document}